\documentclass{article}
\newif\iffinal
\newif\ifrevision

\finaltrue

\ifrevision
\newcommand{\revised}[1]{\textcolor{red}{#1}}
\newcommand{\revisedblock}[1]{\color{red}#1\color{black}}
\else
\newcommand{\revised}[1]{#1}
\newcommand{\revisedblock}[1]{#1}
\fi

\usepackage{arxiv}

\newcommand{\ie}{\emph{i.e.}}
\newcommand{\eg}{\emph{e.g.}}
\newcommand{\etal}{\emph{et al.}}

\usepackage[utf8]{inputenc} % allow utf-8 input
\usepackage[T1]{fontenc}    % use 8-bit T1 fonts
\usepackage{hyperref}       % hyperlinks
\usepackage{url}            % simple URL typesetting
\usepackage{booktabs}       % professional-quality tables
\usepackage{amsfonts}       % blackboard math symbols
\usepackage{amsmath}
\usepackage{nicefrac}       % compact symbols for 1/2, etc.
\usepackage{microtype}      % microtypography
\usepackage{xcolor}         % colors
\usepackage{subcaption}

\usepackage{graphicx}
\usepackage{listings} % for source code typeset
\usepackage{bm}

\title{Bridging Text and Crystal Structures: Literature-driven Contrastive Learning for Materials Science}

\iffinal
\author{%
Yuta Suzuki$^{1}$ \quad Tatsunori Taniai$^{2}$ \quad Ryo Igarashi$^2$ \quad Kotaro Saito$^{3,4}$ \quad Naoya Chiba$^4$ \\
\textbf{Yoshitaka Ushiku}$^2$ \quad \textbf{Kanta Ono}$^4$\thanks{Corresponding author: ono@ap.eng.osaka-u.ac.jp}\\
$^1$Toyota Motor Corporation \quad $^2$OMRON SINIC X Corporation \quad $^3$Randeft, Inc.\\
$^4$The University of Osaka
}
\else
\author{Anonymous Authors\\\\
An Anonymous Submission Under Review\\
}
\fi

\begin{document}
\maketitle

\begin{abstract}
Understanding structure--property relationships is an essential yet challenging aspect of materials discovery and development. To facilitate this process, recent studies in materials informatics have sought latent embedding spaces of crystal structures to capture their similarities based on properties and functionalities. However, abstract feature-based embedding spaces are human-unfriendly and prevent intuitive and efficient exploration of the vast materials space. Here we introduce Contrastive Language--Structure Pre-training (CLaSP), a learning paradigm for constructing crossmodal embedding spaces between crystal structures and texts. CLaSP aims to achieve material embeddings that 1) capture property- and functionality-related similarities between crystal structures and 2) allow intuitive retrieval of materials via user-provided description texts as queries. To compensate for the lack of sufficient datasets linking crystal structures with textual descriptions, CLaSP leverages a dataset of over 400,000 published crystal structures and corresponding publication records, including paper titles and abstracts, for training. We demonstrate the effectiveness of CLaSP through text-based crystal structure screening and embedding space visualization.
\end{abstract}

\section{Introduction}
The properties of materials, ranging from low-level properties such as bandgap and formation energy to high-level functionalities such as superconductivity, are determined by their crystal structures~\cite{callisterMaterialsScienceEngeneering2010, degraefStructureMaterials2012}. Thus, unlocking the structure--property relationships of materials is key to accelerating materials discovery and development.

AI-driven materials science pursues this ambition through the use of machine learning (ML). 
One area of research has focused on predicting material properties using graph neural networks~\cite{xieCrystalGraphConvolutional2018, chenGraphNetworksUniversal2019, chenUniversalGraphDeep2022, lin2023efficient} and transformers~\cite{yan2022periodic, taniai2024crystalformer, ito2025crystalframer}, leveraging large-scale crystal structure datasets annotated with properties simulated by first-principles calculations. Although this approach has shown success, these models are specialized for specific simulatable properties, such as bandgap, and are unable to provide a comprehensive view of materials with diverse properties and functionalities.

Other studies have explored developing embedding spaces for crystal structures to capture their similarities based on properties and functionalities~\cite{xieHierarchicalVisualizationMaterials2018, suzukiSelfsupervisedLearningMaterials2022,liGlobalMappingStructures2023, quLeveragingLanguageRepresentation2024}. However, these efforts are bottlenecked by the lack of dedicated training datasets with diverse property and functionality annotations. Annotating crystal structures is costly, requiring expert knowledge, physical experimentation, or computationally intensive simulations. 
Consequently, these methods often produce abstract, unannotated embedding spaces that are not easily navigable for materials discovery. 
\revised{Because these spaces cannot be queried directly with natural-language descriptions, they remain of limited use to researchers who wish to search by desired properties rather than specific structural identifiers.}

The annotation costs and model interpretability are common problems in ML, leading to the exploration of learning paradigms that use natural language text descriptions, instead of class labels, for supervision. % to learn the correlation between various concepts expressed in the language space and images.
The seminal work, CLIP (Contrastive Language--Image Pre-Training) \cite{radfordLearningTransferableVisual2021}, pioneered this approach by using contrastive learning between image and description text pairs. By learning to align two embedding spaces across the two modalities, CLIP enables crossmodal retrieval between images and texts, and zero-shot recognition of images using text-based prompts.

The success of CLIP has inspired language-supervised representation learning for molecules~\cite{zengDeeplearningSystemBridging2022, wangMultimodalChemicalInformation2022, liuMultimodalMoleculeStructure2023, seidlEnhancingActivityPrediction2023, takedaMultimodalFoundationModel2023, kaufmanCOATIMultimodalContrastive2024} and materials~\cite{moroMultimodalFoundationModels2025, ozawaGraphtextContrastiveLearning}.
However, these methods for materials use textual descriptions about structural features rather than properties~\cite{moroMultimodalFoundationModels2025, ozawaGraphtextContrastiveLearning}, thus limiting their ability to capture high-level information such as material functionalities.

\begin{figure}[t]\centering
\includegraphics[width=150mm]{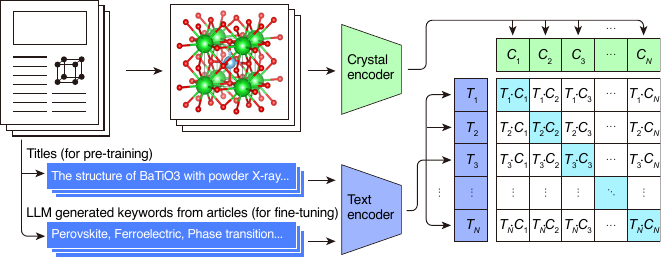}
    \caption{\textbf{Contrastive learning paradigm of CLaSP in two stages.} (1) Pre-training using pairs of crystal structures and publication titles. (2) Fine-tuning using pairs of crystal structures and keywords that are generated from the titles and abstracts using an LLM.}
    \label{fig:clasp_outline}
\end{figure}

To overcome this limitation, we introduce Contrastive Language--Structure Pre-training (CLaSP) (Fig.~\ref{fig:clasp_outline}). CLaSP leverages a large-scale dataset of published crystal structures and corresponding article information retrieved from the Crystallography Open Database (COD)~\cite{Grazulis2009}. Specifically, we utilize publication titles for pre-training and keyword-based captions, generated from pairs of titles and abstracts using a large language model (LLM), for fine-tuning. 
We hypothesize that these textual sources provide a comprehensive representation of material characteristics. By leveraging bibliographic data as natural descriptions of structures, we bypass the need for labor-intensive, specialized annotations, enabling large-scale training of crossmodal models.

\revised{Our central objective is to learn a joint language-structure representation in which crystal structures are organized according to the semantic concepts expressed in human language. In this representation, a user can supply a free‑form textual description (e.g., ``narrow-bandgap material'') and retrieve candidate structures, while the model can simultaneously assign such semantic labels to previously unlabeled structures.}
\revised{
This capability distinguishes our approach from conventional text-based search systems that rely solely on existing textual annotations.
These conventional search systems falter when textual metadata are absent or incomplete for structures, which is frequently the case for newly simulated or experimentally determined structures.}
%Moreover, when textual metadata are absent or incomplete—as is frequently the case for newly simulated or experimentally determined structures—conventional language‑based retrieval approaches falter.
%This practical bottleneck calls for a model that can infer semantic properties directly from crystal structure data while still allowing intuitive language‑based interaction.}

We demonstrate the effectiveness of CLaSP through two key applications: intuitive text-based material retrieval and materials space mapping. Extensive analyses show that CLaSP effectively learns structure embeddings that capture abstract and complex material concepts, such as `superconductor' and `metal-organic frameworks'.

% \revisedblock{Moreover, CLaSP functions as a foundation model for materials science, trained on a vast, diverse corpus of crystal structures and textual descriptions. Leveraging this broad coverage, it can address a wide range of material properties without requiring new data or specialized models, heralding a new paradigm of flexible, scalable ML in materials discovery.}

% \subsection{Contrastive learning using published crystal structures}
% \label{sec:dataset}
%\section{Results}
\section{Contrastive language--structure pre-training}
\label{sec:method}
We propose using a dataset of crystal structures paired with their publication information, such as titles and abstracts, for language--structure contrastive learning. By assuming that these texts convey material characteristics,  this approach aims to link crystal structure embeddings with material properties and functionalities through human-interpretable linguistic semantics.

% 512,312 
We used a total of 406,048 crystal structures associated with paper titles retrieved from the COD, as detailed in Sec.~\ref{sec:data_retrieval}. CLaSP uses these captioned structures to jointly train a crystal encoder and a text encoder. For each training iteration, the crystal encoder transforms a batch of crystal structures into embeddings $\{\bm{c}_i\}$, whereas the text encoder transforms the paired caption texts into embeddings $\{\bm{t}_i\}$.
This training procedure is outlined in Fig.~\ref{fig:clasp_outline}.
CLaSP aligns the two encoders by minimizing the large margin cosine loss function~\cite{wangCosFaceLargeMargin2018}:
\begin{equation}
\label{eq:loss}
L = - \frac{1}{N} \sum_{i=1}^N \log \frac{\exp(s(\cos(\bm{c}_i, \bm{t}_i) - m))}{\exp(s(\cos(\bm{c}_i, \bm{t}_i) - m)) + \sum_{j=1, j \neq i}^N \exp(s \, \cos(\bm{c}_i, \bm{t}_j))},
\end{equation}
% where $N$ is the batch size, and $s > 0$ and $m \in [0, 1]$ are hyper-parameters for scaling and margin, respectively. 
which essentially increases the positive-pair similarities $\cos(\bm{c}_i, \bm{t}_i)$ (\ie, diagonal elements of the affinity matrix in Fig.~\ref{fig:clasp_outline}) while reducing the negative-pair similarities $\cos(\bm{c}_i, \bm{t}_j)$ (\ie, off-diagonal elements).

Here, $N$ is the batch size. The scaling hyperparameter $s > 0$ amplifies cosine similarities to make the loss function more sensitive to small similarity differences, thereby enhancing training effectiveness. The margin hyperparameter $m \in [0, 1]$  enforces a gap between the positive-pair similarities and the negative-pair similarities.
This loss function is equivalent to the cross entropy loss in CLIP~\cite{radfordLearningTransferableVisual2021} when $m = 0$. We found that incorporating the margin leads to better generalization in downstream tasks, as we will show in Sec.~\ref{sec:verification-loss-func}.
% Appendix~\ref{sec:appendix_hyperparameter}.

To learn crystal embeddings $\{\bm{c}_i\}$, a CGCNN model~\cite{xieCrystalGraphConvolutional2018} \revised{was trained from scratch with a redesigned output head. We replaced the original property regression layer with a linear projection that directly produces an embedding vector. Consequently, no property labels (\eg, formation energy) were used or predicted during training; the crystal encoder was supervised solely through the contrastive loss in Eq.~\eqref{eq:loss} so that its embedding aligns with the paired text embedding.} 
% To learn crystal embeddings $\{\bm{c}_i\}$, a CGCNN model~\cite{xieCrystalGraphConvolutional2018} was trained from scratch to serve as the crysta encoder. 
For text embeddings $\{\bm{t}_i\}$, a frozen pre-trained SciBERT model~\cite{Beltagy2019SciBERT} was used as the text encoder, followed by a three-layer multilayer perceptron (MLP) fed with the \texttt{CLS} token embedding. 
We used the embedding dimensionality of 768 for both modalities.
We provide the detailed training procedure in Sec.~\ref{sec:encoder_training_details}.

\begin{figure}[t]
\centering
\includegraphics[width=\textwidth]{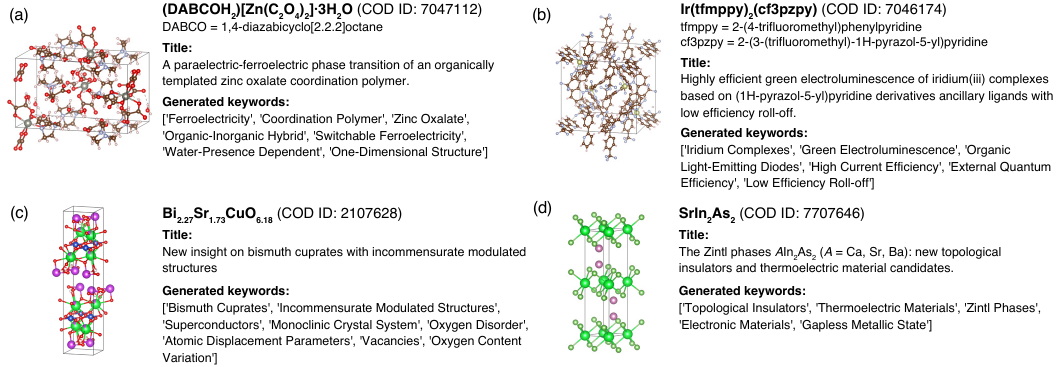}
\caption{\textbf{Example of crystal structures with publication titles and generated keywords.} Panels (a)--(d) showcase dataset entries~\cite{pasinskaParaelectricFerroelectricPhase2018, liPEGCopperIHalide2018, mironovNewInsightBismuth2016, ogunbunmiZintlPhasesAIn2As22021} whose titles or keywords contain `ferroelectric', `electroluminescence', `superconductor', and `thermoelectric', respectively.}
    \label{fig:crystal_caption_example}
\end{figure}

We consider keyword-based crystal structure screening as a demonstrative downstream task and hence perform fine-tuning using keyword captions instead of titles. To this end, we identified abstracts for 80,813 entries of the training set and generated keywords for these entries from their title--abstract pairs using an LLM. We used Meta's Llama~3~\cite{grattafiori2024llama3herdmodels} to generate up to 10 keywords, such as `visible light photocatalysis' and `narrow bandgap', for each crystal structure.
Examples of dataset entries are displayed in Fig.~\ref{fig:crystal_caption_example}, and the overall dataset generation procedure is detailed in Sec.~\ref{sec:data_retrieval}--\ref{sec:keyword_generation}.

% \section{Experiments}
% \label{sec:experiments}

% Discussionの前か後に入れるのもあり
\section{Related work}
\label{sec:related_work}
Our work builds upon several lines of research in materials informatics, ML, and natural language processing (NLP). Below, we briefly review these studies and discuss their relevance and differences with our approach. 

\subsection{Crystal structure modeling}
Crystal structure modeling using neural networks serves as a fundamental basis for ML models that process crystal structures. One of its most straightforward applications is crystal property prediction, for which various crystal encoder architectures have been proposed.
% <A brief historical review of the literature here>
These architectures typically employ geometric graph neural networks (GNNs) that model interatomic interactions via neural message passing. Early successful examples include CGCNN~\cite{xieCrystalGraphConvolutional2018} and SchNet~\cite{schutt18schnet}, which represent crystal structures as distance-based graphs. Subsequent work has aimed to capture richer interaction representations by incorporating bond and global state attributes~\cite{chenGraphNetworksUniversal2019}, three-body interactions~\cite{choudhary21alignn}, attention mechanisms~\cite{louisGraphConvolutionalNeural2020,yan2022periodic,taniai2024crystalformer, ito2025crystalframer}, and infinite periodicity~\cite{lin2023efficient,taniai2024crystalformer}.

%Various graph neural network (GNN) approaches have been proposed for ML models that directly take crystal structures as input. Early successful examples include CGCNN\cite{xieCrystalGraphConvolutional2018} and MEGNet\cite{chenGraphNetworksUniversal2019}, which represent crystal structures as graphs. Subsequent work has aimed to capture richer interaction representations by incorporating angular information and three-body interactions\cite{chenUniversalGraphDeep2022}, attention mechanisms \cite{louisGraphConvolutionalNeural2020} and periodic symmetry\cite{yan2022periodic, lin2023efficient, taniai2024crystalformer}.
 % iCGCNN, ALIGNN, Matformer, PotNet, Crystalformer

These GNN models demonstrate high accuracy in predicting specific properties, such as bandgap and formation energy. However, they all rely on the availability of large-scale datasets (\eg, the Materials Project~\cite{jainCommentaryMaterialsProject2013} and JARVIS-DFT~\cite{choudharyJointAutomatedRepository2020}) with property labels simulated by first-principles calculations for training. As a result, these approaches face challenges when applied to domains where training data are limited or poorly curated. They are also restricted to predicting specific target properties rather than capturing the comprehensive characteristics of materials.

More recent work focuses on developing surrogate models for interatomic potentials~\cite{chenUniversalGraphDeep2022,dengCHGNetPretrainedUniversal2023,batatiaFoundationModelAtomistic2024,yangMatterSimDeepLearning2024}, rather than targeting specific properties. However, like earlier approaches, these models also depend on large-scale training datasets with simulated property labels, such as the Materials Project~\cite{jainCommentaryMaterialsProject2013} or the Open Catalyst Dataset~\cite{chanussotOpenCatalyst20202021, tranOpenCatalyst20222023}.

%These graph-based models have shown high accuracy in predicting specific properties such as bandgaps by leveraging large-scale datasets (\eg, the Materials Project \cite{jainCommentaryMaterialsProject2013} and OQMD \cite{kirklinOpenQuantumMaterials2015}) derived from first-principles calculations. However, they all focus on tasks for which labels are available for the target property or functionality, posing challenges in applying them to domains where such data are absent or insufficiently curated. More recent models such as M3GNet\cite{chenUniversalGraphDeep2022}, MACE\cite{batatiaFoundationModelAtomistic2024}, CHGNet\cite{dengCHGNetPretrainedUniversal2023}, and MatterSim \cite{yangMatterSimDeepLearning2024} serve as surrogate models for interatomic potentials, yet they, too, rely on large-scale training data, such as the Materials Project \cite{jainCommentaryMaterialsProject2013} or the Open Catalyst Dataset \cite{chanussotOpenCatalyst20202021, tranOpenCatalyst20222023}. 

% <Discussion on the relevance and differences with our work here. Include discussion on Figure 3 again. Also mention which encoders are used, and why.>
In our work, we introduce a novel learning paradigm for crystal structure modeling by leveraging language supervision from publication records. This approach enables the model to learn diverse material characteristics---ranging from bandgap to superconductivity---without requiring explicitly labeled training data. 
To demonstrate this approach, we adopt CGCNN~\cite{xieCrystalGraphConvolutional2018} as the crystal encoder for its simplicity and established recognition as a baseline model. The high efficiency of CGCNN is particularly important for our COD dataset, which consists of over 400,000 structures with an average of about 190 atoms per material.

%To demonstrate this approach, we adopt the widely-recognized CGCNN~\cite{xieCrystalGraphConvolutional2018} as the crystal encoder. The high computational efficiency of CGCNN enables scalable training. This is particularly important for our COD dataset, which contains relatively large structures (i.e., with an average of approximately 190 atoms per material). %As CGCNN is a widely recognized baseline architecture, we can establish baseline performance for the demonstration of the proposed approach.

%In this study, we adopt CGCNN\cite{xieCrystalGraphConvolutional2018}, known for its simplicity and established baseline performance, rather than more sophisticated encoders. We make this choice because CGCNN's lightweight architecture is more amenable to repeated training on large datasets, and our focus is on demonstrating the effect of aligning crystal structures with text rather than advancing encoder design.
% Against this background, our study does not aim to develop a new crystal structure encoder but rather focuses on multimodal contrastive learning as a strategy that does not depend on data with rich annotations. In this study, we adopt CGCNN—known for its simplicity and established baseline performance—as our crystal structure encoder to more clearly demonstrate the effects of our training approach.

\subsection{Embedding learning for materials and crystal structures}
\label{sec:related_work_embedding}
Embedding learning is another actively explored application of crystal structure modeling, aimed at understanding the materials space by capturing property- and functionality-level similarities. 
Early studies focused on learning atom representations for limited material classes (\eg, alkali metals and metalloids)~\cite{xieHierarchicalVisualizationMaterials2018,zhouLearningAtomsMaterials2018,ryan2018crystal}. For example, Xie~\etal~\cite{xieHierarchicalVisualizationMaterials2018} derived atom embeddings by utilizing latent features from a GNN trained on a property prediction task. 

To learn more comprehensive material representations on a large scale, 
Suzuki~\etal~\cite{suzukiSelfsupervisedLearningMaterials2022} proposed a global mapping of 120k diverse materials via self-supervised contrastive learning between crystal structures and diffraction patterns. Remarkably, their embedding space reportedly captured complex material concepts, such as superconductors and lithium-ion battery materials, without requiring annotated training data.
Li~\etal~\cite{liGlobalMappingStructures2023} investigated similar global mappings using various material fingerprint representations in a property prediction pre-training approach. 

However, these embedding approaches often lack interpretability for humans, as the meaning of individual dimensions within the learned embedding vectors is difficult to discern. Because of this, Suzuki~\etal~\cite{suzukiSelfsupervisedLearningMaterials2022} relied on a query structure with known properties to retrieve materials with similar properties. This limitation underscores the growing need for more intuitive and interpretable methods to facilitate functionality-driven materials exploration and design.

\revised{To mitigate this limitation}, more recent studies have explored extending material representations through multimodal learning with texts~\cite{moroMultimodalFoundationModels2025,ozawaGraphtextContrastiveLearning}, aiming to attribute semantic meaning to material embeddings.
For example, Ozawa~\etal~\cite{ozawaGraphtextContrastiveLearning} used crystal structures along with rule-based textual descriptions of their geometric features for contrastive learning. 
\revised{While the individual embedding dimensions of these structure embeddings are not meant to be interpreted directly, co-training with text in a shared latent space enables semantic interpretation via proximity to descriptive text embeddings---allowing for semantic inference even without manually labeled structures at deployment time.}
However, these methods rely on textual descriptions of structural features rather than material properties, thus limiting their ability to capture high-level information such as material functionalities.

In our work, we demonstrate that publication records in the materials science literature can effectively teach models about the complex properties of associated structures, by leveraging advances in language modeling discussed below.

\subsection{Language modeling and multimodal learning}
\label{sec:language_modeling_and_multimodal_learning}
Our study leverages language modeling for its three crucial roles in ML applications utilizing textual data: machine understanding, human interpretation, and language supervision.

Historically, a primary goal of ML-based language modeling has been to enable machines to understand textual information. For instance, word2vec~\cite{mikolovEfficientEstimationWord2013} was proposed to learn semantic vector representations of words from their co-occurrences in texts, and seq2seq models~\cite{sutskever2014seq2seq,vaswani2017transformer} enabled context-aware translation and summarization. In particular, the transformer architecture~\cite{vaswani2017transformer} excels at parallel computation and capturing long-range dependencies. 
This innovative architecture has fueled large-scale self-supervised pre-training of language models~\cite{devlin2019bert}, driving significant progress in machine text understanding and ultimately leading to the advent of LLMs with impressive capabilities, such as the GPT~\cite{radford2018gpt1,radford2019gpt2,brownLanguageModelsAre2020,openai2024gpt4technicalreport} and Llama~\cite{touvron2023llamaopenefficientfoundation,touvron2023llama2openfoundation,grattafiori2024llama3herdmodels} series.

An early and notable application of these NLP methods in materials science was demonstrated by Tshitoyan~\etal~\cite{tshitoyanUnsupervisedWordEmbeddings2019}. They applied word2vec~\cite{mikolovEfficientEstimationWord2013} to a corpus of 3.3 million scientific abstracts from the materials science literature. Their findings showed that the learned word embeddings captured latent knowledge about materials science, including novel insights that were not explicitly presented in the used corpus. This suggests the potential for data-driven research through mining the massive body of scientific literature, as well as the informativeness of abstracts.

Our study builds on more recent advances in language modeling, specifically by adopting SciBERT~\cite{Beltagy2019SciBERT} as the text encoder~(Sec.~\ref{sec:method}) and utilizing Llama~3~\cite{grattafiori2024llama3herdmodels} for keyword extraction from title-abstract pairs in dataset generation~(Sec.~\ref{sec:keyword_generation}). 
SciBERT is a variant of BERT~\cite{devlin2019bert} pre-trained on a large corpus of scientific publications, thus particularly suited for analyzing the bibliographic texts.

Since language is one of the most intuitive forms of information for humans, it is also useful to enhance the interpretability and interactability of ML models through multimodal learning~\cite{radfordLearningTransferableVisual2021,bommasaniOpportunitiesRisksFoundation2022}. As discussed in the previous section, our work and other recent studies~\cite{moroMultimodalFoundationModels2025,ozawaGraphtextContrastiveLearning} employ contrastive learning between texts and structures to embed linguistic semantics into structural representations.

Furthermore, when powerful language models are integrated into this multimodal learning paradigm, they can derive strong supervision from noisy natural-language texts. As demonstrated by CLIP~\cite{radfordLearningTransferableVisual2021} and other studies~\cite{jiaScalingVisualVisionLanguage2021,bommasaniOpportunitiesRisksFoundation2022} on image data, this learning strategy treats natural language as a high-freedom annotation, eliminating the need for high-quality labeled datasets. 

A major difference between our work and these computer vision applications~\cite{radfordLearningTransferableVisual2021,jiaScalingVisualVisionLanguage2021} is that, in computer vision, crowd-labeled datasets (e.g., ImageNet~\cite{deng2009imagenet} and MS COCO~\cite{lin2014mscoco}) are often available or affordable. In contrast, materials science lacks large-scale datasets that describe material properties and functionalities. This is primarily because annotating crystal structures requires expert knowledge or labor-intensive experimentation, making crowd-sourcing impractical. Our work addresses this issue by leveraging bibliographic records associated with structures as annotations, inspired by the word2vec approach of Tshitoyan~\etal~\cite{tshitoyanUnsupervisedWordEmbeddings2019}.

\section{Results}
\label{sec:results}
In this section, we demonstrate the effectiveness of the proposed approach both quantitatively and qualitatively through two applications: text-based material retrieval and embedding space visualization. 
For text-based material retrieval (Sec.~\ref{sec:zero-shot-screening}), we evaluate how accurately the model links crystal structures with their textual descriptions, such as `superconductor' and `narrow-bandgap material'. %This evaluation also highlights an advantage over conventional property prediction models. 
Additionally, we verify the design of our loss function through this retrieval application (Sec.~\ref{sec:verification-loss-func}).
For embedding space visualization, we construct an intuitive map of the materials space (Sec.~\ref{sec:embedding-visualization}).
\revised{We also utilize these two applications to compare the proposed method with an existing embedding approach (Sec.~\ref{sec:comparison_existing_approach}).}

\subsection{Zero-shot crystal structure screening by texts}
\label{sec:zero-shot-screening}
To evaluate CLaSP's ability to link crystal structures with textual property descriptions, we performed crystal structure retrieval using keywords representing material functionalities (\emph{e.g.}, `thermoelectric' and `superconductor'). Given the embedding of a query keyword, we retrieved structure embeddings from the test set that showed high cosine similarities with the keyword embedding.
\revised{Once trained, this application requires only a database of unannotated test structures, relying on publication-associated structures solely during training. By enabling semantic inference for previously unseen or unannotated structures, it represents a fundamental departure from conventional text-based retrieval systems.}
%\revised{Although the individual dimensions of these structure embeddings are not intended to be inspected in isolation, co-training them with textual representations in a shared latent space offers a key benefit: each structure embedding can be interpreted via its proximity to text embeddings that describe material characteristics, enabling functionality inference without manually labeled structures.}

\revised{For evaluation, we assessed the retrieval performance of the proposed approach by using publication titles associated with the test materials as hidden labels.}
We regarded a structure to possess a queried property if the associated paper title contained the keyword or its variations  (\eg, given `superconductor' as a query, the terms `superconductive' and `superconductivity' were also considered correct). 
\revised{This evaluation essentially demonstrates how competitively the proposed method performs compared to conventional text-based retrieval, despite the absence of textual annotations for the target structures.}

\revised{Retrieval performance is reported with two complementary metrics.
We examine the trade-off between true- and false-positive rates with the ROC (Receiver Operating Characteristic) curve. The area under this curve (ROC-AUC) equals the probability that a randomly chosen positive sample scores higher than a randomly chosen negative one, thus reflecting the overall ordering among all 40\,604 candidates (test set).
In contrast, Average Precision (AP) condenses the entire precision–recall curve into a single value by integrating precision over all recall levels. Thus, it emphasizes how densely true positives populate the upper part of the ranking---the first few dozen entries that most researchers actually inspect.
Because our target materials account for less than 0.4\% of the dataset (see Table~\ref{tb:test_scores}), the overwhelming surplus of negatives would drive AP toward zero. To keep the AP scores informative, we evaluate AP on a subset of the test set where the negatives are randomly downsampled to match the number of positives.
Overall, ROC-AUC measures the global discriminative power, whereas AP evaluates whether the positives are ranked near the top.
%With positives accounting for less than 0.4 \% of the dataset, ROC-AUC gauges global discriminative power, whereas AP shows whether these scarce positives are surfaced near the top.  
%In such an imbalanced setting, the overwhelming surplus of negatives would otherwise drive AP toward zero.  
%To keep the score informative, we therefore compute AP after retaining all positives and randomly down-sampling an equal number of negatives.
}

\revised{Table~\ref{tb:test_scores} summarizes the ROC-AUC and AP scores for six query keywords representing some functionality-level material concepts, evaluated using pre-trained (PT) and fine-tined (FT) models. Interestingly, even for material concepts with only a few hundred positive samples in the training set (corresponding to 0.05--0.38\% of the training set), the model can still learn to identify various functionalities at a notably high performance level. Given the complexity of these material properties, this is quite remarkable and suggests that the learned representations can generalize effectively under limited supervision.}

\begin{figure}[p]
\centering
    \includegraphics[width=0.98\textwidth]{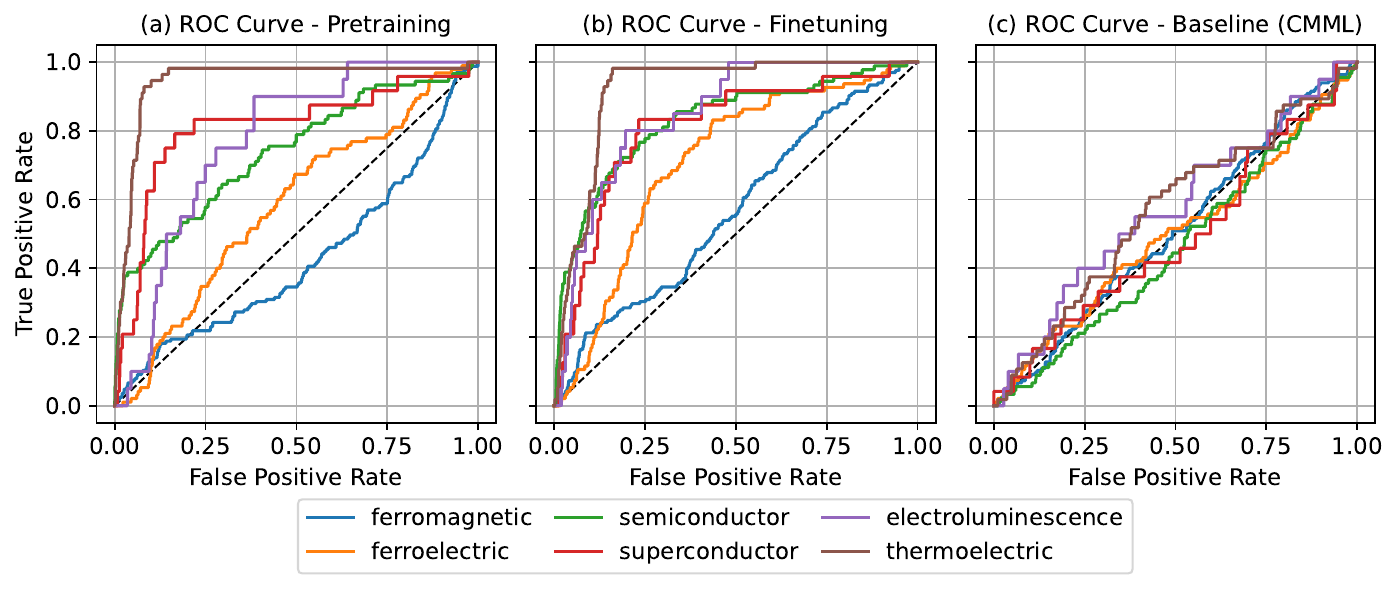}
    \caption{\textbf{ROC curves of keyword-based crystal structure retrieval.} (a) The zero-shot results with only pre-training show good performance, and (b) fine-tuning leads to further improvements. 
    \revised{(c) The baseline method CMML~\cite{suzukiSelfsupervisedLearningMaterials2022}, which is trained solely on crystal structure information and does not utilize any text descriptions, lags behind our proposed approach that leverages both structure and textual data.} 
The test set consists of 40,604 materials, and the dashed diagonal lines represent the ROC curves for random selection.}
    \label{fig:roc_plot}
\end{figure}

Figures~\ref{fig:roc_plot}a and \ref{fig:roc_plot}b show detailed
ROC curves for these six keywords before and after fine-tuning, with dashed diagonal lines representing random selection. 
% For a more comprehensive evaluation,
% Table~\ref{tb:test_scores} summarizes the mean and individual ROC-AUC scores for the six keywords, along with the AP scores.
While the zero-shot prediction using the pre-trained model (Fig.~\ref{fig:roc_plot}a) already demonstrates good performance, with a mean ROC-AUC of 0.7185, fine-tuning (Fig.~\ref{fig:roc_plot}b) further improved it to 0.7804. 
A similar trend is observed in the AP evaluation (Table~\ref{tb:test_scores}).
% \revised{
% In contrast, the baseline method CMML~\cite{suzukiSelfsupervisedLearningMaterials2022} (Fig.~\ref{fig:roc_plot}, right), which is trained solely on crystal structure information and does not utilize any text descriptions, demonstrates lower retrieval performance, highlighting the benefit of our proposed approach that leverages both structural and textual data. 

% A detailed comparison with CMML is provided later (Sec.~\ref{sec:comparison_existing_approach})}.
% These results highlight the ability of the CLaSP model to capture complex structure--property relationships across diverse material functionalities by analyzing crystal structures alone. Section~\ref{sec:discussion_limitations} will discuss these findings in more detail.

%% NOTE: これは今言わなくていい気がする
% \revised{In contrast, the baseline method CMML~\cite{suzukiSelfsupervisedLearningMaterials2022} (Fig.~\ref{fig:roc_plot}, right) is trained solely on crystal structure information and does not use any text descriptions. 
% We observe that its retrieval performance is comparatively lower. 
% This difference highlights the advantage of our approach, which combines structural and textual data; a detailed comparison with CMML is provided later (Sec.~\ref{sec:comparison_existing_approach}).}

To further validate the retrieval performance, we retrieved the top-100 materials using keywords related to bandgap---specifically, `narrow-bandgap material' and `insulator'---and analyzed their bandgap distributions. 
%we analyzed the bandgap distributions of materials that were retrieved by keywords related to bandgap, specifically `narrow-bandgap material' and `insulator.' 
Since the COD does not provide property labels, we predicted bandgaps of materials by using a state-of-the-art property prediction model, Crystalformer~\cite{taniai2024crystalformer}, with pre-trained model weights (specifically the seven-block model trained on the JARVIS-DFT 3D 2021 dataset) provided by the authors.

\begin{figure}[p]
\centering
    \includegraphics[width=0.75\textwidth]{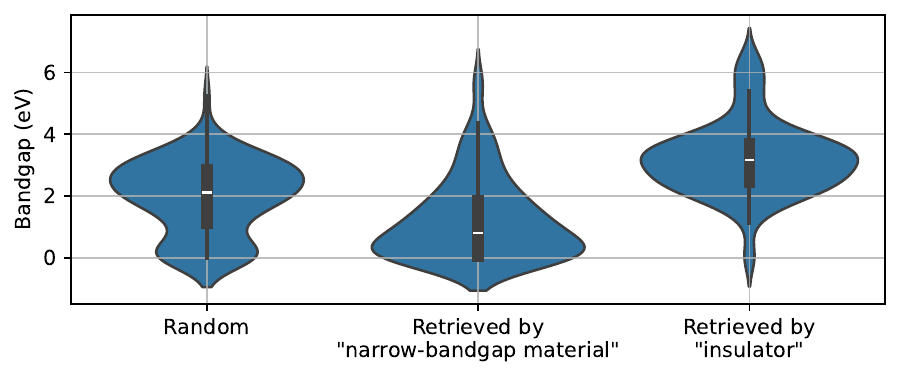}
    \caption{\textbf{Violin plots of bandgaps for crystals retrieved via keyword searches.} The distributions reflect the expected bandgap ranges for narrow bandgap materials and insulators, successfully demonstrating the retrieval of materials with targeted properties.}
    \label{fig:bandgap_violin_plot}
\end{figure}

Figure~\ref{fig:bandgap_violin_plot} shows violin plots of the bandgaps for the retrieved materials. 
These distributions successfully reflect the expected bandgap ranges for narrow-bandgap materials (\ie, bandgaps smaller than $1.1$ eV) and insulators (\ie, large bandgaps), compared to the random sampling distribution.
%See Appendix \ref{sec:appendix_bandgap_prediction} for the t-SNE visualization colored by the predicted bandgap and bandgap prediction details.

% Table1
\begin{table}[p]
    \centering
    \caption{\textbf{ROC-AUC and Average Precision (AP) scores for keyword-based crystal structure retrieval.} ROC-AUC was evaluated on the entire test set (40,604 materials), while AP was evaluated on a balanced subset with randomly subsampled negatives. PT and FT stand for pre-trained and fine-tuned scores, respectively. The numbers in \textbf{bold} indicate the best scores.}
    \label{tb:test_scores}
    \vskip 0.5ex
% \small
    % \begin{tabular}{lccccc}
    % \hline
    %  & & \multicolumn{2}{c}{ROC-AUC} & \multicolumn{2}{c}{AP} \\ \cmidrule(lr){3-4} \cmidrule(lr){5-6}
    % Query & \# Positives & PT & FT & PT & FT \\ \hline
    % ferromagnetic & 165 & 0.4281 & 0.5674 & 0.5018 & 0.5716 \\
    % ferroelectric & 95 & 0.5814 & 0.7097 & 0.5357 & 0.6380 \\
    % semiconductor & 90 & 0.7382 & 0.8290 & 0.7705 & 0.8254 \\
    % electroluminescence & 24 & 0.8106 & 0.8090 & \textbf{0.9050} & 0.8759 \\
    % thermoelectric & 20 & 0.7714 & 0.8496 & 0.7987 & \textbf{0.9124} \\
    % superconductor & 56 & \textbf{0.9431} & \textbf{0.9180} & 0.8880 & 0.8228 \\\hline
    % Avg. & - & 0.7121 & 0.7804 & 0.7333 & 0.7743 \\ \hline
    % \end{tabular}
    % 列指定子: l(Query), r(#Pos Test), r(#Pos Train), r(%Pos Train), c(ROC PT), c(ROC FT), c(AP PT), c(AP FT)
  \begin{tabular}{l r rr c c c c}
  \hline
  & & \multicolumn{2}{c}{\revised{Positives (Train)}} & \multicolumn{2}{c}{ROC-AUC} & \multicolumn{2}{c}{AP} \\
  \cmidrule(lr){3-4} \cmidrule(lr){5-6} \cmidrule(lr){7-8} % booktabs パッケージが必要
  Query & \# Positives (Test) & \# Count & \% Percent & PT & FT & PT & FT \\
  \hline
  ferromagnetic       & 165 & 1241 & 0.38\% & 0.4281 & 0.5674 & 0.5018 & 0.5716 \\
  ferroelectric       & 95  & 746  & 0.23\% & 0.5814 & 0.7097 & 0.5357 & 0.6380 \\
  semiconductor       & 90  & 719  & 0.22\% & 0.7382 & 0.8290 & 0.7705 & 0.8254 \\
  superconductor      & 56  & 465  & 0.14\% & \textbf{0.9431} & \textbf{0.9180} & 0.8880 & 0.8228 \\
  electroluminescence & 24  & 217  & 0.07\% & 0.8106 & 0.8090 & \textbf{0.9050} & 0.8759 \\
  thermoelectric      & 20  & 161  & 0.05\% & 0.7714 & 0.8496 & 0.7987 & \textbf{0.9124} \\
  \hline
  Avg.                & -   & -    & -      & 0.7121 & 0.7804 & 0.7333 & 0.7743 \\
  \hline
  \end{tabular}
\end{table}

\subsection{Verification of loss function design}
\label{sec:verification-loss-func}
Through this keyword-based retrieval task, we also investigated the impact of loss function design (Eq.~\ref{eq:loss}), which was adapted from CosFace~\cite{wangCosFaceLargeMargin2018} instead of CLIP~\cite{radfordLearningTransferableVisual2021}. Specifically, we analyzed how the model's performance depends on the margin and scale hyperparameters in the loss function, coincides with CLIP's loss when the margin is set to zero.
We trained the model with various combinations of margin $m \in \{0, 0.3, 0.5\}$ and scale $s \in \{1.0, 1.5, 2.0, 2.5, 3.0, 3.5\}$, and evaluated the mean ROC-AUC scores both before and after fine-tuning. 

The results in Table~\ref{tb:roc-auc} indicate that both parameters impact performance. Particularly, a higher margin tends to increase validation scores after fine-tuning, suggesting that the margin loss promotes better generalization in downstream tasks. 

% In Appendix~\ref{sec:appendix_clip_results}, we further analyzed the ROC curves of the best CLIP loss model ($m=0$ and $s=2.0$) in Fig.~\ref{fig:roc_plot_clip}, and compared them with  the best CosFace loss model ($m=0.5$ and $s=3.0$) in Fig.~\ref{fig:roc_plot}. The comparison shows that introducing a margin leads to well balanced performance across various keywords. 

\begin{table}[t]
    \centering
    \caption{\textbf{ROC-AUC comparison of keyword-based crystal structure retrieval across different loss hyperparameters.} The numbers in \textbf{bold} indicate the best results and the numbers with \underline{underline} indicate the second best results.}
    \vskip 0.5ex
\label{tb:roc-auc}
% \small
    \begin{tabular}{lcccc}
    \hline
    Loss & Margin  & Scale & Pre-trained (val) & Fine-tuned (val) \\%& Fine-tuned (test) \\
    \hline
    % ### clip s=2.0 のミス修正前 ### 
    % CLIP~\cite{radfordLearningTransferableVisual2021} & 0.0 & 1.0 & 0.6310 & 0.6943 & - \\
    %      & 0.0 & 1.5 & 0.6526 & 0.6521 & - \\
    %      & 0.0 & 2.0 & \textbf{0.7285} & \textbf{0.7837} & \underline{0.7778}\\
    %      & 0.0 & 2.5 & 0.6553 & 0.6791 & - \\
    %      & 0.0 & 3.0 & 0.6946 & 0.7227 & - \\
    %      & 0.0 & 3.5 & 0.6053 & 0.6856 & - \\
    
    % ### 修正後### 
    CLIP~\cite{radfordLearningTransferableVisual2021} & 0.0 & 1.0 & 0.6310 & 0.6943 \\
         & 0.0 & 1.5 & 0.6526 & 0.6521  \\
         & 0.0 & 2.0 & 0.7152 & 0.7336  \\
         & 0.0 & 2.5 & 0.6553 & 0.6791  \\
         & 0.0 & 3.0 & 0.6946 & 0.7227  \\
         & 0.0 & 3.5 & 0.6053 & 0.6856  \\
         \hline
    CosFace~\cite{wangCosFaceLargeMargin2018} & 0.3 & 1.0 & 0.5156 & 0.6495  \\
         & 0.3 & 1.5 & \underline{0.7170} & 0.7273  \\
         & 0.3 & 2.0 & 0.6074 & 0.6701  \\
         & 0.3 & 2.5 & 0.6925 & 0.7365  \\
         & 0.3 & 3.0 & 0.6223 & 0.7395  \\
         & 0.3 & 3.5 & 0.6498 & 0.7496  \\
         & 0.5 & 1.0 & 0.6282 & 0.6994  \\
         & 0.5 & 1.5 & 0.7164 & 0.7763  \\
         & 0.5 & 2.0 & 0.5832 & 0.7006  \\
         & 0.5 & 2.5 & 0.6347 & \underline{0.7778}  \\
         & 0.5 & 3.0 & \textbf{0.7185} & \textbf{0.7828} \\%& 0.7804\\
         & 0.5 & 3.5 & 0.6764 & 0.7031  \\
    \hline
    \end{tabular}
\end{table}

\subsection{Embedding space visualization}
\label{sec:embedding-visualization}
To demonstrate how the proposed language--structure embedding can intuitively navigate the materials space, we created several visualizations of the structure embeddings from the test set using t-SNE.

First, we created a \emph{world map} of COD materials to analyze the alignment of the learned materials space and semantics. % by creating a \emph{world map} of COD materials.
%First, we created a \emph{world map} of COD materials. 
We grouped the structure embeddings into 20 clusters using k-means++ and assigned each cluster an LLM-generated keyword label that summarizes the associated paper titles, as detailed in Sec.~\ref{sec:appendix_method_visualization}.

\begin{figure}[p]\centering
\includegraphics[width=160mm]{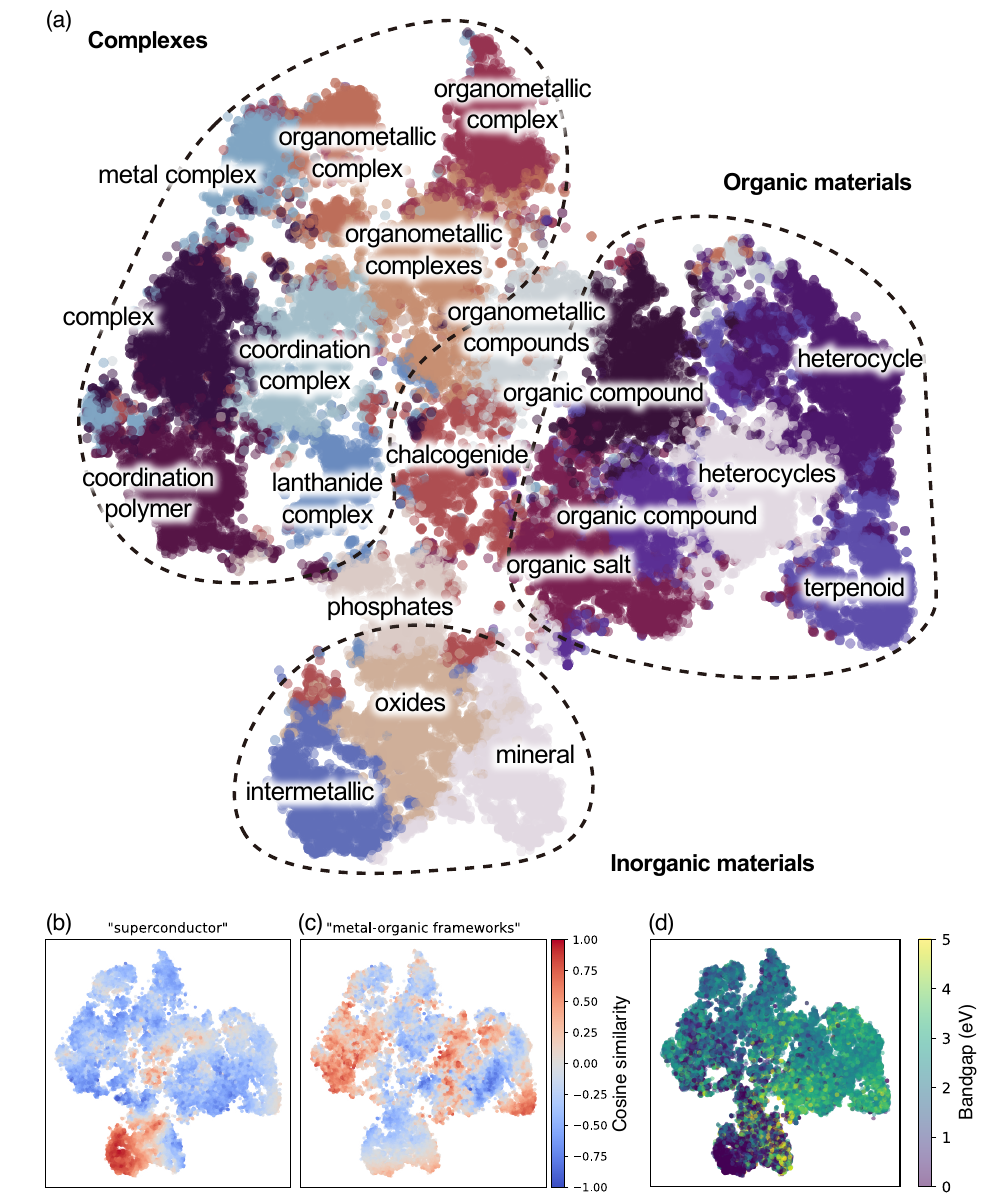}

    \caption{\textbf{t-SNE visualization of crystal structure embeddings.} (a) \emph{World map} of COD materials. The embeddings are grouped into 20 clusters and assigned keywords that represent the paper titles associated with the clusters. (b, c) Heat maps showing cosine similarities between the structure embeddings and query-text embeddings (`superconductor' and `metal-organic frameworks'). (d) Bandgap distribution of crystal structure embeddings. Predicted bandgaps trend to reflect known properties of material clusters in (a), such as larger bandgaps for organic compounds and near-zero bandgaps for intermetallic compounds.}
    \label{fig:crystal_map}
\end{figure}

The resulting map in Fig.~\ref{fig:crystal_map}a shows a meaningful distribution of materials, forming lands of clusters of similar materials, such as an organic materials land, a complexes land, and an inorganic materials land. The map suggests that the model recognizes material similarities that are intuitive to human. 
In contrast, embeddings learned without textual information often fail to capture such high-level semantic relationships, as we will show in the next section.

\begin{figure}[ht]\centering
\includegraphics[width=120mm]{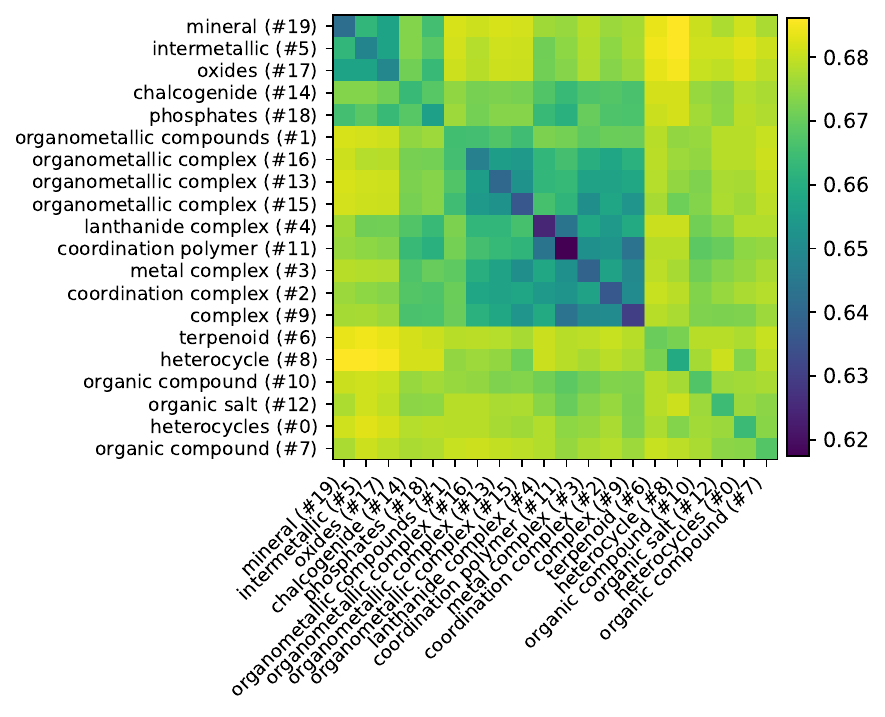}
    \caption{\revised{\textbf{A Jensen--Shannon divergence matrix for cluster coherence and separation analysis.}
    The rows and columns represent the 20 clusters from the map in Fig.~\ref{fig:crystal_map}a. 
    Each matrix element represents the `distance' between two clusters, evaluated based on the symmetric Jensen--Shannon divergence detailed in Sec.~\ref{sec:method_jsd_analysis}. 
    Dark-colored diagonal elements indicate low intra-cluster divergence (\ie, high coherence), while bright-colored off-diagonal elements indicate high inter-cluster divergence (\ie, strong separation).}}
    \label{fig:clustering_JS_heatmap}
\end{figure}

% 追記：JSDを使ったクラスタリング結果の評価
\revisedblock{
The matrix visualization in Fig.~\ref{fig:clustering_JS_heatmap} further quantitatively analyzes the intra-cluster coherence and inter-cluster separation of the map shown in Fig.~\ref{fig:crystal_map}a. Each matrix element represents the `distance' between two clusters, evaluated based on the Jensen--Shannon divergence (JSD) between histograms of words in paper titles (see Sec.~\ref{sec:method_jsd_analysis} for for methodological details).
In a well-formed clustering, the JSD values within clusters should be small, while those between different clusters should be large, resulting in a matrix heatmap with dark diagonal elements.

% To quantitatively analyze the clustering qualities of this map, we analyzed the intra-cluster coherence and inter-cluster separation using a Jensen--Shannon divergence (JSD) matrix (Fig.~\ref{fig:clustering_JS_heatmap}). See Sec.~\ref{sec:method_jsd_analysis} for the method details.
% In a well-formed clustering, the JSD values within clusters should be small, while those between clusters should be large, resulting in a matrix heatmap with a dark diagonal line.

Despite the presence of noise in the JSD between title texts, Fig.~\ref{fig:clustering_JS_heatmap} shows a prominent dark diagonal, indicating strong intra-cluster coherence and clear inter-cluster separation among the title groups.

We also observe two darker square blocks along the diagonal: one in the upper left and the other near the center.
The first block corresponds to clusters dominated by minerals, intermetallic compounds, and oxides, while the second encompasses clusters rich in organometallic and coordination complexes. 
These block-wise proximities reflect the contiguous `lands' seen in Fig.~\ref{fig:crystal_map}, showing that chemically related subfamilies are split into neighboring, finer-grained clusters that nonetheless remain close in the latent space. 
}

% In Fig.~\ref{fig:clustering_JS_heatmap}, the dark diagonal line confirms that the titles grouped within each cluster share similar lexical distributions, signaling internal coherence. Off the diagonal, most cells are much brighter, indicating good separation between clusters that are not chemically related. Superimposed on this pale background are two darker, square blocks on the upper-left and middle. The first covers clusters dominated by minerals, intermetallic compounds and oxides, while the second encompasses clusters rich in organometallic and coordination complexes. Their mutual proximity reproduces the contiguous `lands' seen in Fig.\ref{fig:crystal_map}, showing that chemically related subfamilies are split into neighboring, finer‑grained clusters that nonetheless remain close in the latent space. 
% Between these two continental regions lie intermediary clusters such as chalcogenides and phosphates; their intermediate color reflects their dual affinity for solid‑state and coordination chemistry. 
% % Taken together, the heat‑map pattern demonstrates that the joint language–structure embedding organises materials into clusters that are tight internally yet well separated externally, while faithfully preserving the multilevel hierarchy of chemical relationships present in the dataset.
% }

Furthermore, cosine similarity-based heat maps allow us to easily identify regions relevant to a given text query. Figure~\ref{fig:crystal_map}b shows that `superconductor' is highly correlated with intermetallic compounds and oxides, while Fig.~\ref{fig:crystal_map}c shows `metal-organic frameworks' is aligned with organic compounds.

Finally, we verified the alignment of the map constitution with material properties by overlaying predicted properties onto the map. As done in Sec.~\ref{sec:zero-shot-screening}, we used the pre-trained Crystalformer~\cite{taniai2024crystalformer} to predict the bandgaps of the COD materials.
The resulting bandgap distribution in Fig.~\ref{fig:crystal_map}d shows consistencies with the map (Fig.~\ref{fig:crystal_map}a). For example, the right part in Fig.~\ref{fig:crystal_map}d with larger bandgaps corresponds to organic compounds in Fig.~\ref{fig:crystal_map}a, and the bottom part with near-zero bandgaps corresponds to intermetallic compounds.
These results suggest that the embeddings not only capture intuitive semantics of materials  but also reflect their similarities in terms of material properties.

\begin{figure}[tbh]
  % 最初の図---------------------------
  \centering
  \begin{minipage}[b]{0.8\hsize}
    \centering
    \includegraphics[width=\textwidth]{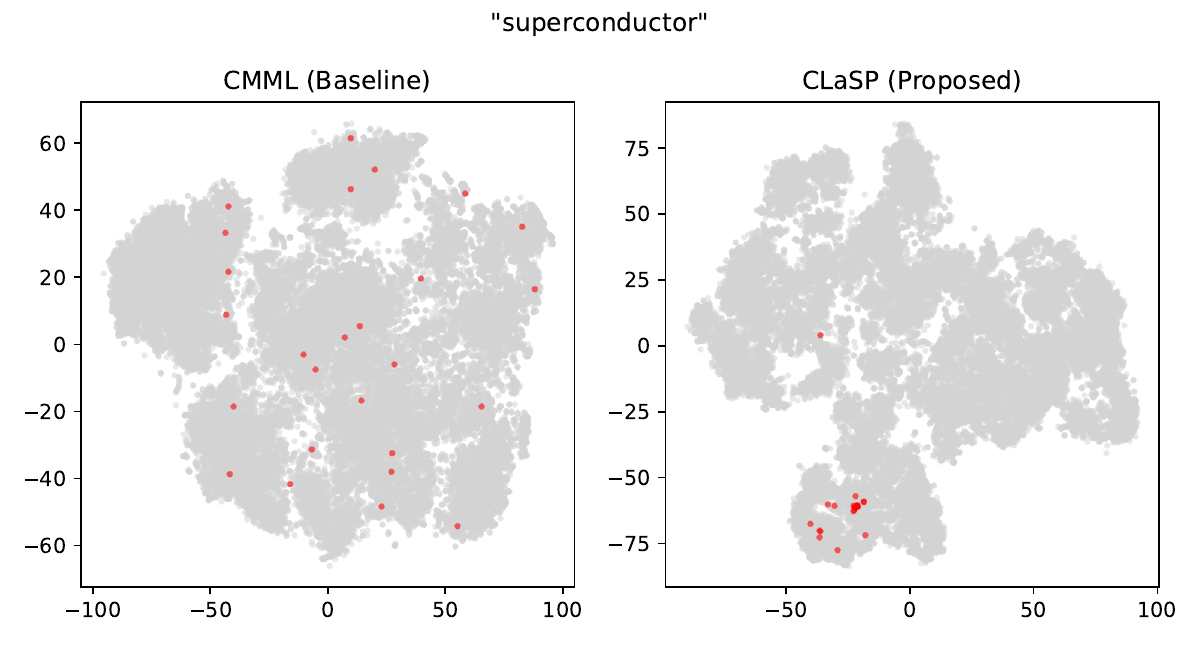}
    % \subcaption{caption1}\label{fig:comparison_clasp_dml_sc}
  \vspace{1mm}
  \end{minipage}\\
  % \vspace{15mm}
  % 2番目の図--------------------------
  \begin{minipage}[b]{0.8\hsize}
    \centering
    \includegraphics[width=\textwidth]{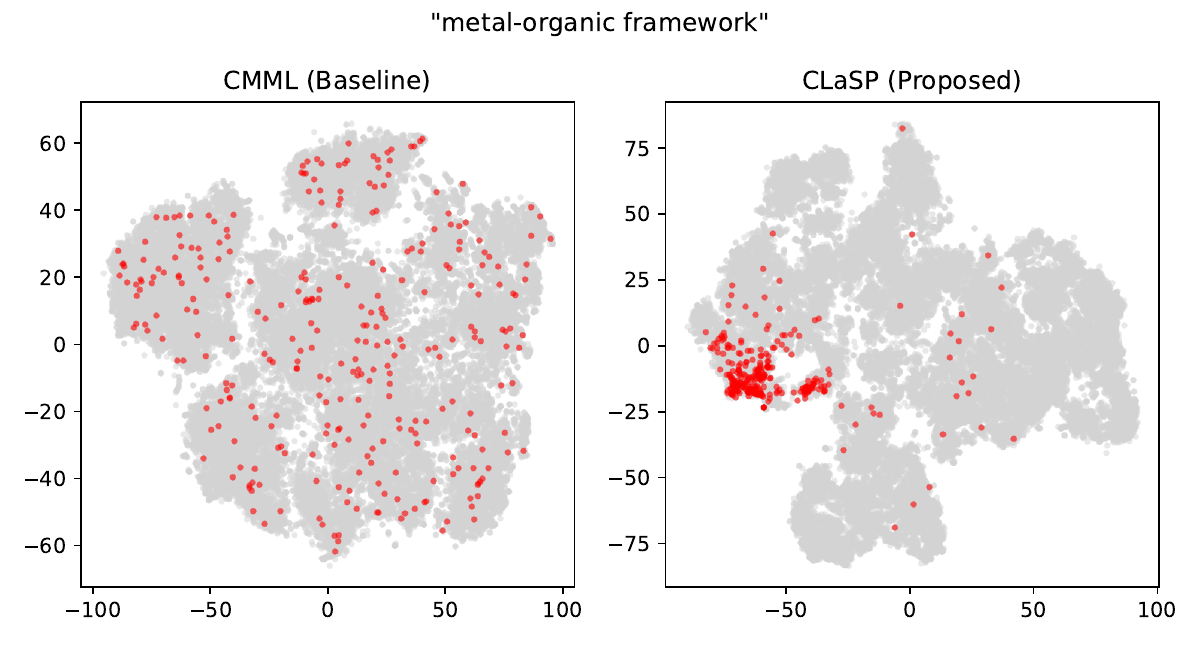}
        % \subcaption{caption2}\label{fig:comparison_clasp_dml_mof}
  \end{minipage}
  \caption{\textbf{t-SNE visualizations of crystal structure embeddings comparing CMML~\cite{suzukiSelfsupervisedLearningMaterials2022} and CLaSP.} Material entries with publication titles that include the keywords `superconductor' (top row) or `metal-organic framework' (bottom row) are highlighted in red.}
  \label{fig:comparison_clasp_dml}
\end{figure}

\subsection{Comparison with an existing approach} \label{sec:comparison_existing_approach}
We compared CLaSP with an existing crystal embedding learning approach called Contrastive Materials Metric Learning (CMML)~\cite{suzukiSelfsupervisedLearningMaterials2022}. CMML also employs contrastive learning, but it aligns the embeddings of two complementary structural representations: crystal structures and their corresponding X-ray diffraction (XRD) patterns. Since XRD patterns can be easily simulated from crystal structures, CMML enables self-supervised learning that only requires a collection of unannotated crystal structures for training. 

\revisedblock{
However, CMML lacks a built-in mechanism for projecting text into its embedding space, thereby preventing direct comparison in text-based retrieval tasks.
We therefore devised a proxy procedure that enables a fair, keyword-based retrieval comparison. For each keyword used in Sec.~\ref{sec:zero-shot-screening}, we collected all COD training structures whose publication titles contained the keyword, averaged their CMML embeddings, and used the resulting vector as a representative concept embedding. Given this vector, retrieval proceeds exactly as with CLaSP: we compute the cosine similarities to the embeddings of the 40,604 COD test structures and rank them accordingly.

Figure~\ref{fig:roc_plot}c shows the ROC curves for CMML on the six keywords, in comparison to the CLaSP results shown in Fig.~\ref{fig:roc_plot}a.
These results suggest that materials semantics cannot be captured without utilizing both structural and textual information.}

% この結果を解釈するために、続いてembedding spaceの視覚的分析を行った。
\revised{To interpret these results, we next visually examined how each model maps semantically similar materials in its embedding space.} 
We created t-SNE visualizations of these crystal embeddings, highlighting entry points whose corresponding publication titles included specific keywords---specifically, `superconductor' and `metal-organic framework.' 

The comparisons in Fig.~\ref{fig:comparison_clasp_dml} show that, while CMML randomly scatters these keyword-specified entries across the map (left), CLaSP highly concentrates them in specific areas in the map (right). 
These concentrated areas also align with the areas of intermetallic compounds and organic compounds in the materials map in Fig.~\ref{fig:crystal_map}a.

These results highlight a key advantage of CLaSP. By incorporating textual information during training, CLaSP learns to recognize similarities between materials based not only on their structures but also on their properties and functionalities through text-based supervision. In contrast, CMML, which relies solely on structural data, struggles to capture these high-level relationships among materials, particularly when they exhibit diverse structures or compositions.

\clearpage
\section{Discussion and limitations}
\label{sec:discussion_limitations}

\subsection{Use of titles and abstracts for supervision}
Both the screening and visualization results in Sec.~\ref{sec:results} have confirmed that publication information provides strong supervision for learning crystal structure embeddings and linking them to material properties.

However, titles in the materials science literature tend to highlight specific, potentially intriguing aspects of the reported materials, rather than offering a comprehensive description. For example, in Fig.~\ref{fig:roc_plot}a and Table~\ref{tb:test_scores} (PT), the ROC performance for the keyword `superconductor' outperforms that for `semiconductor' and other keywords, despite the more complex structure–property relationships involved. We hypothesize that this is due to the frequent co-occurrence of superconductivity in the materials and the term `superconductor' in the titles, whereas `semiconductor' is less prioritized.

A similar issue may explain the relatively low retrieval accuracy for the keyword `ferromagnetic' in Fig.~\ref{fig:roc_plot}a and Table~\ref{tb:test_scores} (PT). Since ferromagnetism is a major characteristic of widely used iron-based materials, paper titles often omit the term `ferromagnetic.' This could introduce noise into the title-based supervision and hinder the learning of this material concept.
In contrast, Fig.~\ref{fig:roc_plot}b and Table~\ref{tb:test_scores} (FT) show that fine-tuning using keywords derived from both titles and abstracts improves the overall retrieval performance. This suggests that abstracts convey more comprehensive information about the materials.

These results suggest potential improvements for CLaSP by incorporating richer training data sources beyond publication titles and abstracts, such as full texts, figures, and tables. Additionally, when papers cite other publications that report specific crystal structures, the citation contexts may provide valuable text descriptions for those structures.

\subsection{Dataset biases}
\begin{figure}[thb]
\centering
\includegraphics[width=0.95\textwidth]{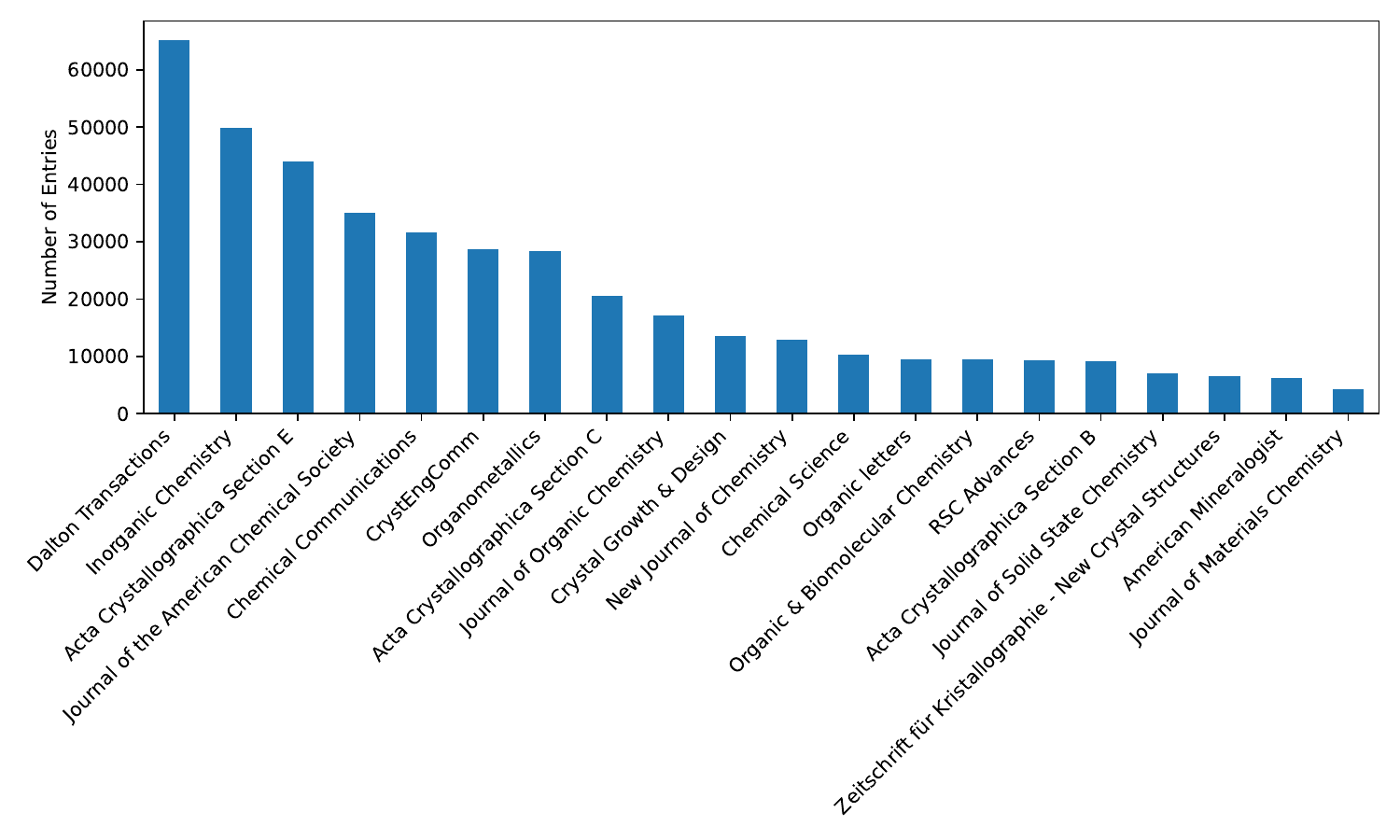}
    \caption{\textbf{Top 20 journals contributing to the COD dataset.}}
    \label{fig:cod_source_journal}
\end{figure}

Our analysis also revealed a potential bias in the COD dataset, which is predominantly composed of crystallography and chemistry publications. Over 80\% of the entries come from a small subset of journals primarily focused on these fields (Fig.~\ref{fig:cod_source_journal}). This bias is understandable, given the COD's historical development and its emphasis on crystallography. However, it highlights the need for more diverse data sources to ensure a comprehensive representation of materials and their properties. 
% \revised{To address this limitation, a scalable, non–labor‑intensive pipeline for provide more informative text captions for crystal structures is essential.}
\revised{A promising alternative to the COD is the Inorganic Crystal Structure Database (ICSD), although it is distributed under a paid license and would prevent us from publicly releasing any resulting dataset if used.}
Given the limited availability of large materials databases with publication records beyond the COD and ICSD, augmentation with external sources, such as citation contexts and Wikipedia entries, could be beneficial. This approach is similar to retrieval-augmented generation (RAG)~\cite{lewis2020rag} used in LLM applications. 
We leave such extensions for future work.

% 谷合さんコメント：
% metal とか oxide とかは概念的に広すぎて（=他の性質と同時に持つ場合がありすぎて）、ユークリッド空間内に、他の性質の島を作りながらmetal/oxideらを一つの大陸として作るというのが、幾何学的に困難で、各島にmetal 領域、oxide領域 みたいな感じで分散している可能性もありますね。そうすると、検索で"metal"の1点からの距離をみてもうまくいかない可能性はありそう。
% 例えば、ferromagnetic metal みたいに、他の何かと条件付けしたりすると、マシになる可能性があるかも。

% テキストで材料の多面的な物性を示す難しさは、オリジナルのCLIPにおいても議論されている。
% プロンプトのテンプレートを変えたりアンサンブルしたり、他の様々なテクニックについても適用が考えられる

% \revised{We therefore intend to extend CLaSP to other structure--literature
% datasets such as the Inorganic Crystal Structure Database~\cite{belskyNewDevelopmentsInorganic2002},
% which covers a broader spectrum of inorganic compounds.
% In the longer term we envisage constructing a unified dataset by
% harvesting crystal structures and associated articles from additional
% journal archives and pre-print servers.
% Such diversification would mitigate the crystallography-centric skew of
% the current COD-based training set and should further improve the
% generalisability of the learned representations.}

\revised{We also acknowledge a potential bias introduced by the LLM-generated
keywords that we used for fine-tuning, as these keywords have not been formally
validated by domain experts.
Nevertheless, their use has demonstrably improved the performance of keyword-based materials retrieval (cf. Fig.~\ref{fig:roc_plot}a vs. \ref{fig:roc_plot}b), which suggests that they possess a reasonable degree of semantic validity.  A systematic
expert evaluation of the generated keywords, along with iterative
refinement of the prompting strategy used for keyword generation,
remains an important direction for future work.}

\section{Conclusions and broader impacts}
\label{sec:conclusion}
In this study, we introduced CLaSP, a literature-driven learning paradigm for constructing crossmodal embedding spaces that connect crystal structures with their textual property descriptions. We demonstrated its effectiveness in learning structure embeddings that capture functionality-level material similarities and in enhancing the materials space with intuitive linguistic semantics. % for efficient navigation.

% These promising results indicate the potential to transform how we explore the vast materials space. Potential applications include crystal structure screening and tagging via text prompts. Furthermore, inspired by text-to-image generation~\cite{rameshHierarchicalTextConditionalImage2022}, CLaSP's text embeddings could guide crystal structure generation models~\cite{jiaoCrystalStructurePredictio2023,zeni2024mattergengenerativemodelinorganic}, enabling innovative applications such as text-to-crystal generation. 
% These advancements promise a more intuitive and efficient approach to exploring the materials space.

\revisedblock{
A fundamental finding of this study is that bibliographic metadata associated with crystal structures can effectively guide the learning of a crossmodal latent space that aligns structures with their intuitive textual descriptions. 
This was successfully demonstrated using simple publication metadata, specifically titles and abstracts.
Given that the contrastive learning paradigm of CLaSP can flexibly incorporate text--structure pairs from any source, a natural and ambitious extension is to harness full-text materials science literature---a rich repository where decades of materials knowledge have been systematically articulated.
In the era of data scaling laws~\cite{xia2024shearedllamaacceleratinglanguage}, the field of materials science increasingly confronts the challenge posed by the scarcity of richly annotated materials data.
The proposed literature-driven learning thus offers a realistic and scalable path forward for data-driven materials discovery.

% Because the contrastive objective relies solely on generic structure--text pairs, CLaSP is inherently dataset‑agnostic and can adopt any dataset without modifying the architecture, much like prior large‑scale or domain‑specific adaptations of CLIP~\cite{jiaScalingVisualVisionLanguage2021, stevensBioCLIPVisionFoundation2024}.
% Using crystal structures accompanied only by bibliographic metadata such as titles and abstracts, which were never intended to serve as detailed captions, CLaSP demonstrates, in the spirit of ALIGN~\cite{jiaScalingVisualVisionLanguage2021}, that large volumes of noisy data still support the learning of a meaningful joint embedding of language and structure, offering a realistic route for materials science where richly labelled data are scarce.
% % 
% Because CLaSP is trained through a flexible text–structure contrastive objective, the approach can readily draw on richer sources such as full‑text articles, promising deeper integration of the accumulated knowledge of materials science in future work.

The resulting crossmodal embedding not only underpins the retrieval and open‑vocabulary classification shown here, but also paves the way for automatic captioning of crystal structures and for text‑conditioned generation of novel candidates, echoing the evolution of vision–language models~\cite{liBLIP2BootstrappingLanguageImage2023, rameshHierarchicalTextConditionalImage2022}.  Taken together, these capabilities promise a more intuitive and efficient exploration of the vast materials design space.}  

%also suggests that indirect supervision could unlock the potential of previously untapped data such as raw experimental data and unannotated images, alleviating the limitation of data scarcity in material science research.

\section{Method details}
\label{sec:methods}
Below, we provide detailed procedures for model training, dataset creation, embedding visualization, and \revised{clustering quality analysis}.

\subsection{Training settings}
\label{sec:encoder_training_details}
We optimized the loss function in Eq.~\ref{eq:loss} with scaling factor $s$ of $3$ and margin $m$ of $0.5$, using stochastic gradient descent with global batch size $N$ equal to 16,384 (2,048 $\times$ 8 GPUs).
Title-based pre-training was performed for a total of 2000 epochs, followed by keyword-based fine-tuning for additional 50 epochs.
We used the AdamW optimizer~\cite{kingmaAdamMethodStochastic2015,loshchilovDecoupledWeightDecay2019} with a constant learning rate of $2 \times 10^{-5}$ for pre-training and $1 \times 10^{-6}$ for fine-tuning.
The networks and training code were implemented in Python using PyTorch~\cite{PyTorch_ansel_etal} and PyTorch Geometric~\cite{Fey_Lenssen_2019_PyG}.
Training was performed on a single server with eight NVIDIA A100 GPUs (80GB VRAM), taking approximately 16 hours overall.

%\subsection{Dataset preparation}
\subsection{Data retrieval}
\label{sec:data_retrieval}
This study used the Crystallography Open Database (COD)~\cite{Grazulis2009} as the source of crystal structure data. Compared to other crystal structure databases, the COD is particularly well-suited for our purposes, as it provides publication information (including titles and DOIs) for each crystal structure entry and is available in the public domain.
% 512,312

From the COD, we retrieved 512,312 pairs of crystal structures and their corresponding publication records as of March 2024. Using the DOIs from these records, we further extracted the abstracts of the papers via the Crossref API. This process collected abstracts for 141,311 entries, representing approximately 27.6\% of the entire dataset.

% 本研究では、結晶構造データベースであるCOD (Crystallography Open Database) をデータソースとして用いた。CODは、他の結晶構造データベースと比較して、構造情報に加えて、論文情報（タイトル、DOI）が提供されている点が本研究の目的に合致している。
% CODから、結晶構造データとその構造を報告した論文のタイトルとDOIのペアをxx件（約50万件）取得した。取得したデータに対して、CrossRef APIを用いて、論文のDOIに基づき各論文のアブストラクトを収集した。その結果、約10万件（約20\%）の論文についてアブストラクトを取得できた。

\subsection{Data preprocessing and spliting}
\label{sec:data_preprocessing}
We filtered out the entries with structures containing more than 500 atomic sites, resulting in a dataset of 406,048 crystal structures with corresponding paper titles and DOIs.
We randomly split the dataset into training, validation, and test sets in an 8:1:1 ratio, yielding 324,838 entries for training, 40,604 for validation, and 40,606 for testing.

We used the train set for title-based pre-training, the validation set for selecting a model checkpoint during pre-training or fine-tuning, and the test set for evaluating the retrieval performance (Figs.~\ref{fig:roc_plot} and  \ref{fig:bandgap_violin_plot}; Table~\ref{tb:test_scores}) and for visualizing the embedding space (Figs.~\ref{fig:crystal_map} and \ref{fig:comparison_clasp_dml}). 
We also used this dataset to generate the keyword-captioned dataset for fine-tuning, as explained next.

% まず、収集したデータから、機械学習モデルの開発に用いるdev setを構築するため、超伝導体と熱電材料の2つの材料特性に注目して少量のデータを抽出した。論文タイトルに "superconductor" または "superconductivity" を含むものを正例(positive sample)、これらのキーワードを含まないものを負例(negative sample)として、それぞれ50件ずつランダムに抽出した。同様に、"thermoelectric" または "thermoelectricity" をキーワードとして、正例と負例をそれぞれ50件ずつ抽出した。
% 次に、dev setを除いた残りのデータに対して、原子サイト数が500を超える巨大な構造を持つデータをフィルタリングした後、train, validation, testに8:1:1の比率でランダムに分割した。最終的なデータセットサイズは、train: 324,838件、validation: 40,604件、test: 40,606件となった。

\subsection{Keyword dataset generation for fine-tuning}
\label{sec:keyword_generation}
% fine-tuning用データセットの個数内訳
% train_dataset:  80813
% val_dataset:  10134
% test_dataset:  10197
We derived the keyword-captioned dataset from the main dataset by removing entries without abstract from each split, ensuring no mixing across the splits. For each entry with a title and abstract pair, we generated up to 10 representative keywords using an LLM, specifically Meta's Llama~3 (70B Instruct)~\cite{grattafiori2024llama3herdmodels}. The prompt template used for keyword generation is listed below. 

The keyword generation process took 36 hours using a server equipped with eight NVIDIA A100 GPUs (80GB VRAM) and an efficient LLM inference framework, vLLM~\cite{kwon2023efficient}. 
Finally, we removed generated keywords if they were unrelated to material properties, such as `crystal structure', `X-ray diffraction', `neutron diffraction', `powder diffraction', and `single-crystal X-ray diffraction'. Entries without any remaining keywords were also removed from the dataset. 
\revised{Although this keyword generation step is time-consuming, it is performed only once in this study.}

This process resulted in 80,813 entries for the training set, which was used to fine-tune the pre-trained model for the keyword-based retrieval task. The remaining two sets, containing 10,134 entries for validation and 10,197 for testing, were never used in this study. Note that the validation during fine-tuning was done based on the mean ROC-AUC score, instead of the validation loss, for the validation set of the main dataset. 

% The keyword generation with title and abstract pair of 141,311 papers took 36 hours using a server equipped with eight NVIDIA A100 GPUs (80GB VRAM) and vLLM~\cite{kwon2023efficient}, efficient inference framework. 
% The prompt template for keyword generation is below. Keywords unrelated to material functions and properties, such as 'Crystal Structure', 'X-ray diffraction', 'Neutron Diffraction', 'Powder Diffraction', and 'Single-Crystal X-ray Diffraction', were removed from the caption. Entries lacking any keywords were also removed from the dataset.

{
\scriptsize
\begin{lstlisting}
def prompt_format_func(material_id, title, abstract):
    return \
    f"""Below are title-abstract pairs for materials science papers dealing with crystal structures. For each paper, list up to 10 keywords in English that describe the features, functions, or applications of the material discussed. Focus on the material itself, and do not include general terms or measurement techniques (e.g., Crystal Structure, Crystal Lattice, X-ray diffraction, Neutron Diffraction, Powder Diffraction). Return the results in json format with the following schema.

    **Example input 1:**
    ```
    ID: 0001
    Title: Enhancement of Critical Temperature in Layered Copper Oxide Superconductors via Lattice Compression Techniques
    Abstract: Superconductivity in copper oxides (cuprates) offers vast potential for technological applications due to their high critical temperatures (Tc). Our research presents a novel approach to enhance Tc in layered cuprate materials through the controlled application of lattice compression. Using advanced crystallographic methods, we systematically altered the interlayer spacing and analyzed the resultant changes in electronic properties. Our findings demonstrate a significant improvement in superconducting behavior at elevated temperatures, further supporting the unconventional mechanisms underpinning superconductivity in these materials. 
    ```
    
    **Example output 1:**
    ```json
    [{
        "ID": "0001",
        "Keywords": ["High-Tc", "Cuprate Superconductors", "Lattice Compression", "Electronic Properties", "Layered Structures", "Superconducting Phase", "Temperature Enhancement", "Unconventional Superconductivity"]
    }]
    ```

    **Example input 2:**
    ```
    ID: 0002
    Title: Advancements in Biodegradable Polymers for Sustained Drug Delivery Systems
    Abstract: The development of biocompatible and biodegradable materials is critical in the field of medical implants and drug delivery systems. This paper examines the latest advancements in biodegradable polymers tailored for sustained release of therapeutic agents. We analyze various polymer compositions that provide controlled degradation rates and compatibility with a range of drugs. Our results show promising applications in long-term treatments, reducing the need for repeated administration and improving patient compliance.
    ```
    
    **Example output 2:**
    ```json
    [{
        "ID": "0002",
        "Keywords": ["Biomaterials", "Biodegradable Polymers", "Sustained Release", "Drug Delivery Systems", "Biocompatibility", "Controlled Degradation", "Therapeutic Agents", "Medical Implants", "Long-Term Treatment"]
    }]
    ```

    **Input :**
    ```
    ID: {material_id}
    Title: {title}
    Abstract: {abstract} 
    ```

    **Output :**
    ```json
    """
\end{lstlisting}
}

\subsection{Visualizations} \label{sec:appendix_method_visualization}
Crystal structures in Figs.~\ref{fig:clasp_outline} and \ref{fig:crystal_caption_example} were visualized using VESTA~\cite{mommaVESTAThreedimensionalVisualization2011}.
For the embedding space visualizations shown in Figs.~\ref{fig:crystal_map} and \ref{fig:comparison_clasp_dml}, we employed the t-SNE algorithm~\cite{maatenVisualizingDataUsing2008} , as implemented in openTSNE~\cite{Policar2024openTSNE}. In Fig.\ref{fig:crystal_map}, embedding clusters were generated using the k-means++ algorithm~\cite{arthurKmeansAdvantagesCareful}. These clusters were further annotated by summarizing the associated paper titles using Google Gemini 1.5 Pro with a default temperature parameter of 1.0.

\revisedblock{
\subsection{Clustering quality analysis}
\label{sec:method_jsd_analysis}
Let each article title be represented by an L1-normalized TF-IDF (term frequency-inverse document frequency)~\cite{sparckjonesStatisticalInterpretationTerm1972} vector, denoted  as $\bm{p}_k$. 
Each $\bm{p}_k$ can thus be interpreted as a histogram of the words in the title, normalized to form a probability distribution. The TF-IDF weighting scheme assigns lower weights to common, uninformative words such as ``the'' and ``and'' by incorporating their inverse frequency across the entire test set corpus.

Given a cluster $C_i$, we define its centroid as
\begin{equation}
\bm{\mu}_i = \frac{1}{|C_i|}\sum_{k\in C_i} \bm{p}_k.    
\end{equation}
We then compute the average Jensen–Shannon divergence between each cluster centroid and the entries in another cluster, resulting in the following matrix:
\begin{equation}
M^{\mathrm{JS}}_{ij} = \frac{1}{|C_j|}\sum_{k\in C_j} JS\bigl(\bm{\mu}_i,\,\bm{p}_k\bigr).
\end{equation}
The visualization in Fig.~\ref{fig:clustering_JS_heatmap} is based on the following symmetrized matrix:
\begin{equation}
M^{\mathrm{SJS}}_{ij} = \frac{1}{2}\!\left( M^{\mathrm{JS}}_{ij}+M^{\mathrm{JS}}_{ji}\right).
\end{equation}

Intuitively, the diagonal elements of $M^{\mathrm{SJS}}_{jj}$ quantify intra-cluster coherence, with smaller values indicating tighter clusters.
In contrast, the off-diagonal elements reflect inter-cluster separation, with larger values indicating better separation.
% ちゃんと解釈の式を書かない場合は、余計なことを言わないほうが良さそう
%the centroid-to-centroid distance combined with cluster spread, with larger values indicating better separation.
}

\iffinal
\section*{Acknowledgments}
%This work was primarily supported by JST-Mirai Program Grant Number JPMJMI21G2.
T.T. is partly supported by JSPS KAKENHI Grant Number 24K23911.
R.I. is partly supported by JSPS KAKENHI Grant Number 24K23910.
N.C. is partly supported by JSPS KAKENHI Grant Number 21K14130.
Y.U. is partly supported by JST-Mirai Program Grant Number JPMJMI21G2 and JST Moonshot R\&D Program Grant Number JPMJMS2236.
K.O. is partly supported by the MEXT Program: Data Creation and Utilization-Type Material Research and Development Project (Digital Transformation Initiative Center for Magnetic Materials) Grant Number JPMXP1122715503.
\fi

\section*{Data availability statement}
This study used the Crystallography Open Database (COD)~\cite{Grazulis2009} as the source of crystal structure data. Please see Sec.~\ref{sec:data_retrieval}--\ref{sec:keyword_generation} for the data retrieval and processing details.
\iffinal
%\section*{Code availability}
%We will release our code upon acceptance of this work for publication.
All code for dataset creation, model implementation, training, and analysis, along with pretrained model weights and the resulting datasets, is available at: \url{https://github.com/Toyota/clasp} .
% (URL is provided in the final published version).

\else
%\section*{Code availability}
% The source code that supports the findings and trained model weights are available at \url{https://github.com/quantumbeam/clasp}. (\textit{Note for reviewers: This repository is not yet available. Please refer to supplementary materials for code.})
\fi

\iffinal
\section*{Author contributions}
%YS conceived the concept, developed the model, conducted the analysis, and drafted the manuscript.
\textbf{Y.S.} conceived the concept, developed the model, conducted the analysis, and drafted both the initial manuscript and the revision.
% TT provided guidance on the methodology, analysis and writing, assisted with the literature review, and revised the manuscript draft.
\textbf{T.T.} provided guidance on the methodology, analysis and writing; assisted with the literature review; revised the manuscript draft; and led the review response and revision.
\textbf{R.I.} and \textbf{K.S.} contributed expertise in materials science, advising on the analysis.
\textbf{N.C.} contributed expertise in machine learning, advising on the methodology.
\textbf{Y.U.} co-led materials-related collaborations, provided guidance on the plan and methodology from a machine learning perspective, and advised on the writing.
\textbf{K.O.} co-led materials-related collaborations and provided guidance on the plan and analysis from a materials science perspective.
\fi

\section*{Conflict of interest}
The author declares no competing interests.

% \section*{References}
%\bibliographystyle{naturemag.bst}
%\bibliographystyle{nature}
%\bibliographystyle{bibstyles/hacm.bst}
\bibliographystyle{naturemag-doi-eprint.bst}
\bibliography{references}

\end{document}